\documentclass[runningheads]{llncs}
\usepackage{graphicx}
\DeclareGraphicsExtensions{.pdf,.png}

\usepackage[cmex10]{amsmath}
\usepackage{amssymb}
\usepackage{booktabs}
\usepackage[caption = false,font = footnotesize]{subfig}

\newcommand{\vect}[1]{\mathbf{#1}}
\DeclareMathSymbol{\R}{\mathalpha}{AMSb}{"52}

\begin{document}
\title{Multikernel activation functions: formulation and a case study\thanks{The work of S. Scardapane was supported in part by Italian MIUR, ``\textit{Progetti di Ricerca di Rilevante Interesse Nazionale}'',  GAUChO project, under Grant 2015YPXH4W\_004.}}
%
%
\author{Simone Scardapane\inst{1}\orcidID{0000-0003-0881-8344} \and
Elena Nieddu\inst{2} \and
Donatella Firmani\inst{2}\orcidID{0000-0003-0358-3208} \and
Paolo Merialdo\inst{2}\orcidID{0000-0002-3852-8092}}
\authorrunning{S. Scardapane et al.}
%
\institute{DIET Department, Sapienza University of Rome, Rome, Italy
\email{\{simone.scardapane\}@uniroma1.it} \and
Department of Engineering, Roma Tre University, Rome, Italy\\
\email{firstname.lastname@uniroma3.com}}
\maketitle              
\begin{abstract}
The design of activation functions is a growing research area in the field of neural networks. In particular, instead of using fixed point-wise functions (e.g., the rectified linear unit), several authors have proposed ways of learning these functions directly from the data in a non-parametric fashion. In this paper we focus on the kernel activation function (KAF), a recently proposed framework wherein each function is modeled as a one-dimensional kernel model, whose weights are adapted through standard backpropagation-based optimization. One drawback of KAFs is the need to select a single kernel function and its eventual hyper-parameters. To partially overcome this problem, we motivate an extension of the KAF model, in which multiple kernels are linearly combined at every neuron, inspired by the literature on multiple kernel learning. We provide an application of the resulting \textit{multi-KAF} on a realistic use case, specifically handwritten Latin OCR, on a large dataset collected in the context of the `In Codice Ratio' project. Results show that multi-KAFs can improve the accuracy of the convolutional networks previously developed for the task, with faster convergence, even with a smaller number of overall parameters.
\end{abstract}
\section{Introduction}
The recent successes in deep learning owe much to concurrent advancements in the design of activation functions, especially the introduction of the rectified linear unit (ReLU) \cite{glorot2011deep}, and its several variants \cite{he2015delving,clevert2016fast,jin2016deep}. Albeit the vast majority of applications consider fixed activation functions, a growing trend of research lately has focused on designing \textit{flexible} ones, that are able to adapt their shape from the data through the use of additional trainable parameters. Common examples of this are parametric versions of ReLU \cite{he2015delving}, or the more recent beta-Swish function \cite{ramachandran2017swish}. 

In the more extreme case, multiple authors have advocated for \textit{non-parametric} formulations, in which the overall flexibility and number of parameters can be chosen freely by the user on the basis of one or more hyper-parameters. As a result, the trained functions can potentially approximate a much larger family of shapes. Different proposals, however, differ on the way in which each function is modeled, resulting in vastly different characteristics in terms of approximation, optimization, and simplicity of implementation. Examples of non-parametric activation functions are the maxout neuron \cite{goodfellow2013maxout,zhang2014improving}, defined as the maximum over a fixed number of affine functions of its input, the Fourier activation function \cite{eisenach2017nonparametrically}, defined as a linear combination of a predetermined trigonometric basis expansion, or the Hermitian-based expansion \cite{siniscalchi2017adaptation}. We refer to \cite{scardapane2018kafnets} for a fuller overview on the topic.

In this paper we focus on the recently proposed kernel activation function (KAF) \cite{scardapane2018kafnets}, in which each (scalar) function is modeled as a one-dimensional kernel expansion, with the linear mixing coefficients adapted together with all the other parameters of the network during optimization. In \cite{scardapane2018kafnets} it was shown that KAFs can greatly simplify the design of neural networks, allowing to reach higher accuracies, sometimes with a smaller number of hidden layers. Linking neural networks with kernel methods also allows to leverage a large body of literature on the learning of kernel functions (e.g., kernel filters \cite{liu2011kernel}), particularly with respect to their approximation capabilities. At the same time, compared to ReLUs, KAFs introduce a number of additional design choices, most notably the selection of which kernel function to use, and its eventual hyper-parameters (e.g., the bandwidth of the Gaussian kernel). Although in \cite{scardapane2018kafnets} and successive works we mostly focused on the Gaussian kernel, it is not guaranteed to be the optimal one in all applications.

\subsubsection{Contribution of the paper}
To solve the kernel selection problem of KAFs, in this paper we propose an extension inspired to the theory of multiple kernel learning \cite{gonen2011multiple,aiolli2015easymkl}. In the proposed \textit{multi-KAF}, different kernels are linearly combined for every neuron through an additional set of mixing coefficients, adapted during training. In this way, the optimal kernel for each neuron (or a specific mixture of them) can be learned in a principled way during the optimization process. In addition, since in our KAF implementation the points where the kernels are evaluated are fixed, a large amount of computation can be shared between the different kernels, leading to a very small computational overhead overall, as we show in the following sections.

\subsubsection{Case study: In Codice Ratio}
To show the usefulness of the proposed activation functions, we provide a realistic use case by applying them to the data from the `In Codice Ratio' (ICR) project \cite{firmani2018towards}, whose aim is the automatic transcription of a large part of the Vatican Secret Archive.\footnote{\url{http://www.archiviosegretovaticano.va/}} A key component of the project is an OCR tool applied to characters from a Latin handwritten text (see Section \ref{sec:icr} for additional details). In \cite{firmani2017codice} we presented a convolutional neural network (CNN) for this task, which we applied to a dataset of $23$ different Latin characters extracted from a sample selection of pages from the Vatican Register. In this paper we show that, using multi-KAFs, we can increase the accuracy of the CNN even while reducing the number of filters per layer.

\subsubsection{Organization of the paper}
In Section \ref{sec:preliminaries} and \ref{sec:multikafs} we describe standard activation functions for neural networks, KAFs \cite{scardapane2018kafnets}, and the proposed multi-KAFs. The dataset from the ICR project is described in Section \ref{sec:icr}. We then perform a set of experiments in Section \ref{sec:experiments}, before concluding in Section \ref{sec:conclusions}.

\section{Preliminaries}
\label{sec:preliminaries}
\subsection{Feedforward neural networks}

Consider a generic feedforward NN layer, taking as input a vector $\vect{x} \in \R^d$ and producing in output a vector $\vect{y} \in \R^c$:
\begin{equation}
\vect{y} =  g \left( \vect{W}\vect{x} + \vect{b} \right) \,,
\end{equation}
where $\left\{ \vect{W}, \vect{b} \right\}$ are adaptable weight matrices, and $g(\cdot)$ is an element-wise activation function. Multiple layers can be stacked to obtain a complete NN. In the following we focus especially on the choice of $g(\cdot)$, but we note that everything extends immediately to more complex types of layer, including convolutional layers (wherein the matrix product is replaced by a convolutional operator), or recurrent layers \cite{goodfellow2016deep}.

Generally speaking, the activation functions $g(\cdot)$ for the hidden (not last) layers are chosen as simple operations, such as the rectified linear unit (ReLU), originally introduced in \cite{glorot2011deep}:
\begin{equation}
g(s) = \max\left\{ 0, s \right\} \,,
\label{eq:relu}
\end{equation}
where we use the letter $s$ to denote a generic scalar input to the function, i.e., a single \textit{activation} value. An NN is a generic composition of $L$ such layers, denoted as $f(\vect{x})$, that is trained with a dataset of $N$ training samples $\left\{\vect{x}^i, \vect{d}^i\right\}_{i=1}^N$. In the experimental section, in particular, we deal with multi-class classification with $C$ classes, where the desired output $\vect{d}^i$ represents a one-hot encoding of the target class. We train it by minimizing a regularized cross-entropy cost:
\begin{equation}
\min \left\{ - \sum_{i=1}^N \sum_{c=1}^C d^i_c \log\left( f_c(\vect{x}^i \right) + \lambda \cdot \lVert \vect{w} \rVert^2 \right\} \,,
\label{eq:regularizes_loss_function}
\end{equation}
where $\vect{w}$ is a weight vector collecting all the adaptable weights of the network, $\lambda$ is a positive scalar, and we use a subscript to denote the $c$-th element of a vector.

\subsection{Kernel activation functions}
Differently from \eqref{eq:relu}, a KAF can be adapted from the data. In particular, each activation function is modeled in terms of $D$ expansions of the activation $s$ with a kernel function $\kappa$:
\begin{equation}
g(s) = \sum_{i=1}^D \alpha_i \kappa\left(s, d_i\right) \,,
\label{eq:kaf}
\end{equation}
where the scalars $d_i$ form the so-called dictionary, while the scalars $\alpha_i$ are called the mixing coefficients. To make the training problem simpler, the dictionary is fixed beforehand (and not adapted) by sampling $D$ values from the real line, uniformly around zero, and it is shared across the network, while a different set of mixing coefficients is adapted for every neuron. This makes implementation extremely efficient. The integer $D$ is the key hyper-parameter of the model: a higher value of $D$ increases the overall flexibility of each function, at the expense of adding additional mixing coefficients to be adapted.

Any kernel function from the literature can be used in \eqref{eq:kaf}, provided it respects the semi-definiteness property:
\begin{equation}
\sum_{i=1}^D \sum_{j=1}^D \alpha_i \alpha_j \kappa\left(d_i, d_j\right) \ge 0 \,,
\label{eq:psd_kernel}
\end{equation}
for any choice of the mixing coefficients and the dictionary. In practice, \cite{scardapane2018kafnets} and all subsequent papers only used the one-dimensional Gaussian kernel defined as:
\begin{equation}
\kappa(s, d_i) = \exp\left\{-\gamma\left(s - d_i\right)^2\right\} \,,
\label{eq:gaussian_kernel}
\end{equation}
where $\gamma > 0$ is a parameter of the kernel. The value of $\gamma$ influences the `locality' of each $\alpha_i$ with respect to the dictionary. In \cite{scardapane2018kafnets} we proposed the following rule-of-thumb, found empirically:
\begin{equation}
\gamma = \frac{1}{6\Delta^2} \,,
\label{eq:sigma_rule_of_thumb}
\end{equation}
where $\Delta$ is the distance between any two dictionary elements. However, note that neither the Gaussian kernel in \eqref{eq:gaussian_kernel} nor the rule-of-thumb in \eqref{eq:sigma_rule_of_thumb} are optimal in general. For example, simple smooth shapes (like slowly varying polynomials) could be more easily modeled via different types of kernels or much larger values of $\gamma$ with a possibly smaller $D$. These are common problems also in the kernel literature. Leveraging it, in the next section we propose an extension of KAF to mitigate both problems.

\section{Proposed multiple kernel activation functions}
\label{sec:multikafs}

In order to mitigate the problems mentioned in the previous section, assume to have available a set of $M$ \textit{candidate} kernel functions $\kappa_1, \ldots, \kappa_M$. These can be entirely different functions or the same kernel with different choices of its parameters. There has been a vast research on how to successfully combine different kernels to obtain a new one, going under the name of multiple kernel learning (MKL) \cite{aiolli2015easymkl}. For the purpose of this paper we adopt a simple approach, in order to evaluate its feasibility. In particular, we build each KAF with a new kernel given by a linearly weighted sum of the constituents (base) kernels:
\begin{equation}
g(s) = \sum_{i=1}^D \alpha_i \left[ \sum_{m=1}^M \mu_m \kappa_m\left(s, d_i\right) \right] = \sum_{i=1}^D \alpha_i \widetilde{\kappa}\left(s, d_i\right) \,,
\label{eq:proposed_multikaf}
\end{equation}
where $\left\{\mu_m\right\}_{m=1}^M$ are an additional set of mixing coefficients. From the properties of reproducing kernel Hilbert spaces, it is straightforward to show that $\widetilde{\kappa}$ is a valid kernel if its constituents are also valid kernel functions. We call the resulting activation functions \textit{multi-KAFs}. Note that such an approach only introduces $M-1$ additional parameters for each neuron, where $M$ is generally small. In addition, since in our implementation the dictionary is fixed, all kernels are evaluated on the same points, which can greatly simplify the implementation and allows to share a large part of the computation.

\begin{figure}
\centering
\includegraphics[scale=0.7]{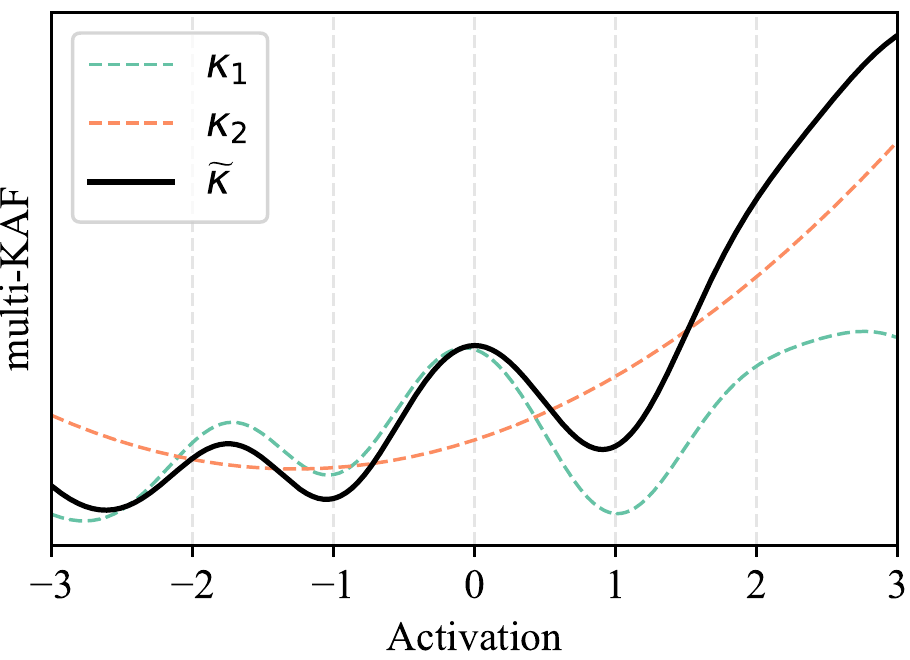}
\caption{An example of multi-KAF obtained from the mixture of a Gaussian kernel (green dashed line) and a polynomial kernel (red dashed line). In the example, we have $D=15$ elements in the dictionary equispaced in $\left[-3.0, 3.0\right]$, mixing coefficients $\alpha_i$ sampled from the normal distribution, and $\mu_1 = 0.5$, $\mu_2 = 0.01$.}
\label{fig:multikaf_example}
\end{figure}

More in detail, for our experiments we consider an implementation with $M=3$, where $\kappa_1$ is the Gaussian kernel in \eqref{eq:gaussian_kernel}, $\kappa_2$ is chosen as the (isotropic) rational quadratic \cite{genton2001classes}:
\begin{equation}
\kappa_2(s, d_i) = 1 + \frac{(s-d_i)^2}{(s-d_i)^2 + c}
\end{equation}
with $c$ being a parameter, and $\kappa_3$ is chosen as the polynomial kernel of order 2:
\begin{equation}
\kappa_3(s, d_i) = \left(1+sd_i\right)^2 \,.
\end{equation}
The rational quadratic is similar to the Gaussian, but it is sometimes preferred in practical applications \cite{genton2001classes}, while the polynomial kernel allows to introduce a smoothly varying global trend to the function. We show a simple example of the mix of $\kappa_1$ and $\kappa_3$ in Fig. \ref{fig:multikaf_example}.

In order to simplify optimization, especially for deeper architectures, we apply the kernel ridge regression initialization procedure described in \cite{scardapane2018kafnets} to initialize all multi-KAFs to a known activation function, i.e., the exponential linear unit (ELU) \cite{clevert2016fast}. For this purpose, denote by $\vect{t}$ the vector of ELU values computed on our dictionary points. We initialize all $\mu_j$ to $\frac{1}{3}$, and initialize the vector of mixing coefficients $\boldsymbol{\alpha}$ as:
\begin{equation}
\boldsymbol{\alpha} = \left(\widetilde{\vect{K}} + \varepsilon\vect{I}\right)^{-1}\vect{t} \,,
\label{eq:kaf_initialization_krr}
\end{equation}
where $\widetilde{\vect{K}} \in \R^{D \times D}$ is the kernel matrix computed between $\vect{t}$ and $\vect{d}$ using $\widetilde{\kappa}$, and $\varepsilon = 10^{-4}$. As a final remark, note that in \eqref{eq:proposed_multikaf} we considered an unrestricted linear combination of the constituting kernels. We can easily obtain more restricted formulations (which are sometimes found in the MKL literature \cite{gonen2011multiple}) by applying some nonlinear transformation to the mixing coefficients $\mu_m$, e.g., a softmax function to obtain convex combinations. We leave such comparisons to a future work.

\section{Case Study: In Codice Ratio}
\label{sec:icr}

As stated in the introduction, we apply the proposed multi-KAF on a realistic case study taken from the ICR project. Apart from the details described below, we refer the reader to \cite{firmani2017codice,firmani2018towards} for a fuller description of the project.

\begin{figure}
\subfloat[]{
\includegraphics[width=0.15\columnwidth,keepaspectratio]{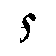}
\label{fig:Immagine1}
} \hfil
\subfloat[]{
\includegraphics[width=0.15\columnwidth,keepaspectratio]{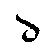}
\label{fig:Immagine2}
} \hfil
\subfloat[]{
\includegraphics[width=0.15\columnwidth,keepaspectratio]{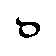}
\label{fig:Immagine3}
} \hfil
\subfloat[]{
\includegraphics[width=0.15\columnwidth,keepaspectratio]{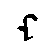}
\label{fig:Immagine4}
} \hfil
\subfloat[]{
\includegraphics[width=0.15\columnwidth,keepaspectratio]{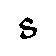}
\label{fig:Immagine5}
} \hfil
\caption{Examples taken from the Latin OCR dataset. (a) and (d) are examples of the character `s'; (b) and (c) are examples of character `d'; (e) is an example of a different version for the character `s' (considered as a separate class in the dataset).}
\label{fig:dataset_examples}
\end{figure}

The overall goal of ICR is the transcription of a large portion of the Vatican Secret Archives, one of the largest existing historical libraries. In the first phase of the project, we collected and manually annotated a set of $23000$ images representing $23$ different types of characters (one of which is a special `non-character' class). All characters were extracted from a sample of $30$ (handwritten) pages of private correspondence of Pope Honorii III from the XIII century. Each character was then annotated using a crowdsourcing platform and $120$ volunteer students.\footnote{The dataset is available on the web at \url{http://www.dia.uniroma3.it/db/icr/}.} A few examples taken from the dataset are shown in Fig. \ref{fig:dataset_examples}.

In \cite{firmani2017codice} we described the design and evaluation of a CNN for tackling the problem of automatically assigning a class to each character, represented as a $56 \times 56$ black-and-white image. The final CNN had the following architecture: (i) a convolutive block with $42$ filters of size $5\times5$; (ii) max-pooling with size $2 \times 2$; (iii) two additional series of convolutive blocks and max-pooling, this time with $28$ filters per layer; (iv) a fully connected layer with $100$ neurons, followed by (v) an output layer of $23$ neurons. In the original implementation, all linear operations were preceded by dropout \cite{srivastava2014dropout} (with probability $50\%$), and followed by ReLU nonlinearities. We will use this dataset and architecture as baseline for our experiments. Note that this design was heavily fine-tuned, and it was not possible to increase the testing accuracy by simply adding more neurons / layers to the resulting CNN model.

\section{Experimental results}
\label{sec:experiments}

\subsection{Experimental setup}
For our comparisons, we use the same CNN architecture described in Section \ref{sec:icr}, but we replace the ReLU functions by either KAFs or the proposed multi-KAFs, with the hyper-parameters described in the previous sections. In order to make the networks comparable in terms of parameters, for KAF and multi-KAF we decrease the number of filters and neurons in the linear layers by $10\%$. To stabilize training, we also replace dropout with a batch normalization step \cite{ioffe2015batch} before applying KAF-based nonlinearities.

We train the networks following a similar procedure as \cite{firmani2017codice}. We use the Adam optimization algorithm on random mini-batches of $32$ elements, with a small regularization factor $\lambda = 0.001$. After every $10$ iterations of the optimization algorithm we evaluate the accuracy on a randomly held-out set of $2500$ samples, taken from the original training set. Training is stopped whenever the validation accuracy has not improved for at least $250$ iterations. The networks are then evaluated on a second independent held-out set of $2300$ examples. All networks are implemented in PyTorch, and experiments are run on a CUDA backend using the Google Colaboratory platform.

\subsection{Experimental results}
\begin{table}[t]
\caption{Results of comparing the different activation functions on the OCR dataset described in Section \ref{sec:icr}. See the text for details on the experimental procedure.}
{\centering\hfill{}
	\setlength{\tabcolsep}{4pt}
	\renewcommand{\arraystretch}{1.5}
	\begin{footnotesize}
	\begin{tabular}{lcc}   
	\toprule
	\textbf{Activation function} & \textbf{Testing accuracy [\%]} & \textbf{Trainable parameters}\\ 
	\midrule
	ReLU & $94.99 \, (\pm 0.1)$ & $191871$ \\
	KAF & $94.00 \, (\pm 0.3)$ & $158840$ \\
	Proposed multi-KAF & $96.31 \, (\pm 0.2)$ & $159552$ \\
	\bottomrule
	\end{tabular}
	\end{footnotesize}
}
\hfill{}
\label{tab:results}
\end{table}

The results of the experiments, averaged over $5$ different repetitions, are shown in Table \ref{tab:results}, together with the number of trainable parameters of the different architectures. It can be seen that, while KAF fails to provide a meaningful improvement in this case, the multi-KAF architecture obtains a significant gain in testing accuracy, stably throughout the repetitions. Most notably, this gain is obtained with a strong decrease in the number of trainable parameters, which is around $16 \cdot 10^{4}$ for the KAF-based architectures, compared to $\approx 19 \cdot 10^{4}$ parameters for the baseline one.

\begin{figure}[!h]
\centering
\subfloat[Training loss]{
\hspace*{-2em}\includegraphics[width=0.60\columnwidth,keepaspectratio]{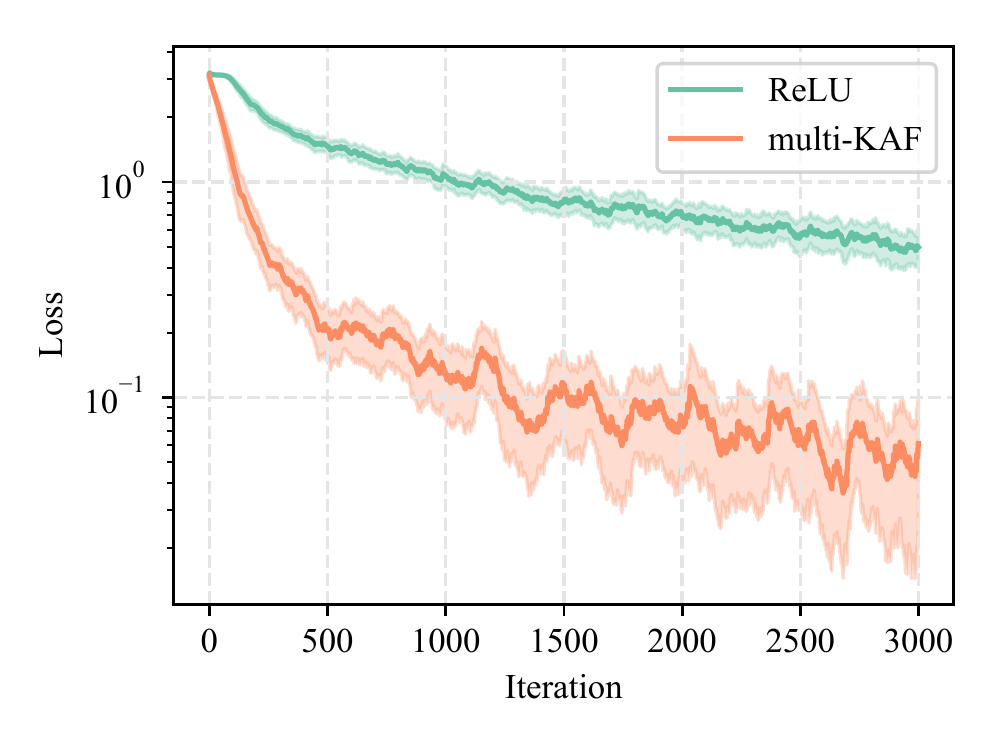}
\label{fig:loss_latin_ocr}
} \vfill
\subfloat[Validation accuracy]{
\hspace*{-2em}\includegraphics[width=0.60\columnwidth,keepaspectratio]{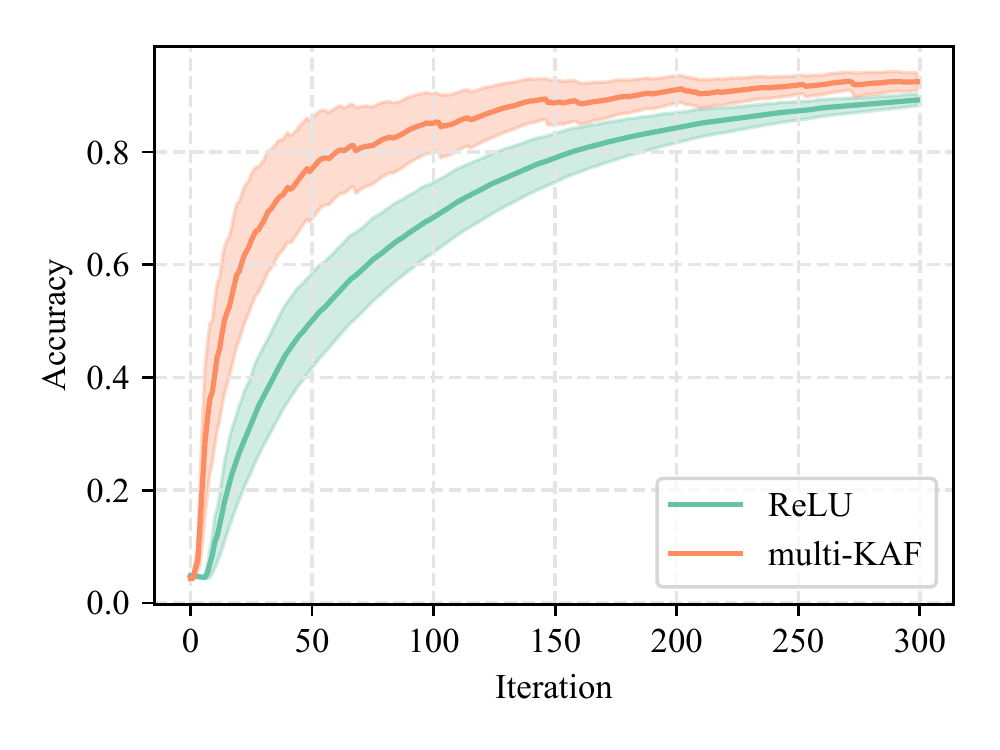}
\label{fig:accuracy_latin_ocr}
} \hfil
\caption{Loss and validation accuracy evolution for the baseline network (using ReLUs) and the proposed multi-KAF. Standard deviation for all curves is shown with a lighter color.}
\label{fig:loss_and_accuracy_latin_ocr}
\end{figure}

This gain is accuracy is not only obtained with a smaller number of overall parameters, but also with a much faster rate of convergence. To see this, we plot in Fig. \ref{fig:loss_and_accuracy_latin_ocr} the evolution of the loss function in \eqref{eq:regularizes_loss_function} and the evolution of the accuracy on the validation portion of the dataset.

\begin{figure}[!t]
\centering
\hspace*{-8em}\includegraphics[scale=0.4]{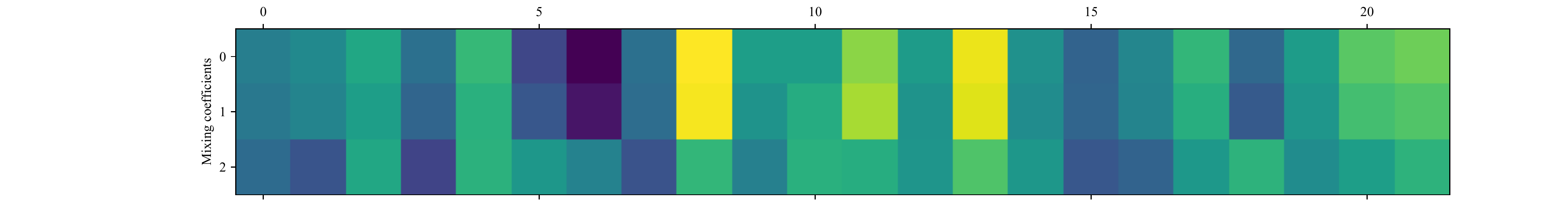}
\caption{Mixing coefficients of the multiple kernel after training, for the first convolutive layer and $25$ randomly sampled filters. Light colors indicate stronger mixing coefficients.}
\label{fig:nu}
\end{figure}

Finally, we also show the final mixing coefficients $\mu_m$ for $25$ randomly sampled neurons of the first convolutive layer of the multi-KAF network, after training, in Fig. \ref{fig:nu}. Interestingly, different neurons require vastly different combinations of base kernels, reflected by light/dark colors in Fig. \ref{fig:nu}.

\section{Conclusions}
\label{sec:conclusions}
In this paper we investigated a new non-parametric activation function for neural networks, which extends the recently proposed kernel activation function, by incorporating ideas from the field of multiple kernel learning to simplify the choice of the kernel function and further increase the expressiveness.

We evaluated the resulting multi-KAF on a benchmark dataset of Latin handwritten characters recognition, in the context of an ongoing real-world project. While in a sense these are only preliminary results, they point to the greater flexibility of such activation functions, coming with a faster rate of convergence and an overall smaller number of trainable parameters for the full architecture. We are currently in the process of collecting a larger dataset, considering a bigger amount of possible classes for the characters, in order to further evaluate the proposed architecture.

Additional research directions will consider the application of multi-KAFs on different types of benchmarks, going beyond standard CNNs, particularly with respect to recurrent models, complex-valued kernels \cite{scardapane2018complex}, and generative adversarial networks. Furthermore, we plan to test more extensively additional types of kernels, such as the periodic ones \cite{genton2001classes}, in order to evaluate the scalability of multi-KAFs in the presence of a larger number of constituent kernels. Generalization bounds for the architecture are also a promising research direction.

%
%
%
\bibliographystyle{splncs04}
\bibliography{biblio}
\end{document}